\documentclass[10pt,twocolumn,letterpaper]{article}
\pdfoutput=1

\usepackage{iccv}              
\usepackage{graphicx}
\usepackage{amsmath}
\usepackage{amssymb}
\usepackage{booktabs}
\usepackage{multirow}
\usepackage{booktabs, multirow} 
\usepackage{soul}
\usepackage{changepage,threeparttable} 
\usepackage{wrapfig}
\usepackage{soul}

\setstcolor{red}
\newcommand{\paragraphskip}{\vspace{0.15cm}}
\newcommand{\red}[1]{{\color{Maroon}#1}}

\newcommand{\green}[1]{\textcolor{OliveGreen}{#1}}

\definecolor{iccvblue}{rgb}{0.21,0.49,0.74}
\definecolor{urlred}{rgb}{0.85, 0.2, 0.55}
\usepackage[pagebackref,breaklinks,colorlinks,allcolors=iccvblue,urlcolor=urlred]{hyperref}
\def\paperID{7367} 
\def\confName{ICCV}
\def\confYear{2025}

\title{Temporal Overlapping Prediction: A Self-supervised Pre-training Method for LiDAR Moving Object Segmentation}

\author{
Ziliang Miao$^{1}$ \quad
Runjian Chen$^{1}$\thanks{Project lead.} \quad
Yixi Cai$^{2}$ \quad
Buwei He$^{2}$ \quad
\\
Wenquan Zhao$^{3}$ \quad
Wenqi Shao$^{4}$ \quad
Bo Zhang$^{4}$ \quad
Fu Zhang$^{1}$\thanks{Corresponding author.}
\\[3mm]
$^1$The University of Hong Kong \quad
$^2$KTH Royal Institute of Technology \quad
\\
$^3$Southern University of Science and Technology \quad
$^4$Shanghai AI Laboratory \\
{\tt\small \{miaozl, rjchen\}@connect.hku.hk \quad \{yixica, buwei\}@kth.se} \\
{\tt\small zhaowq2021@sustech.edu.cn \quad \{shaowenqi,zhangbo\}@pjlab.org.cn\quad fuzhang@hku.hk}}

\begin{document}
\maketitle
\begin{abstract}
    Moving object segmentation (MOS) on LiDAR point clouds is crucial for autonomous systems such as self-driving vehicles. While previous supervised approaches rely on costly manual annotations, LiDAR sequences naturally capture temporal motion cues that can be leveraged for self-supervised learning. 
    In this paper, we propose \textbf{T}emporal \textbf{O}verlapping \textbf{P}rediction (\textbf{TOP}), a self-supervised pre-training method designed to alleviate this annotation burden. \textbf{TOP} learns powerful spatiotemporal representations by predicting the occupancy states of temporal overlapping points that are commonly observed in current and adjacent scans.
    To further ground these representations in the current scene's geometry, we introduce an auxiliary pre-training objective of reconstructing the occupancy of the current scan.
    Extensive experiments on the nuScenes and SemanticKITTI datasets validate our method's effectiveness. \textbf{TOP} consistently outperforms existing supervised and self-supervised pre-training baselines across both point-level Intersection-over-Union (IoU) and object-level Recall metrics.
    Notably, it achieves a relative improvement of up to 28.77\% over a training-from-scratch baseline and demonstrates strong transferability across LiDAR setups.
    Our code is publicly available at \url{https://github.com/ZiliangMiao/TOP}.
\end{abstract}
\section{Introduction}
\label{sec:intro}
\begin{figure*}[ht]
    \centering
    \includegraphics[width=0.8\linewidth]{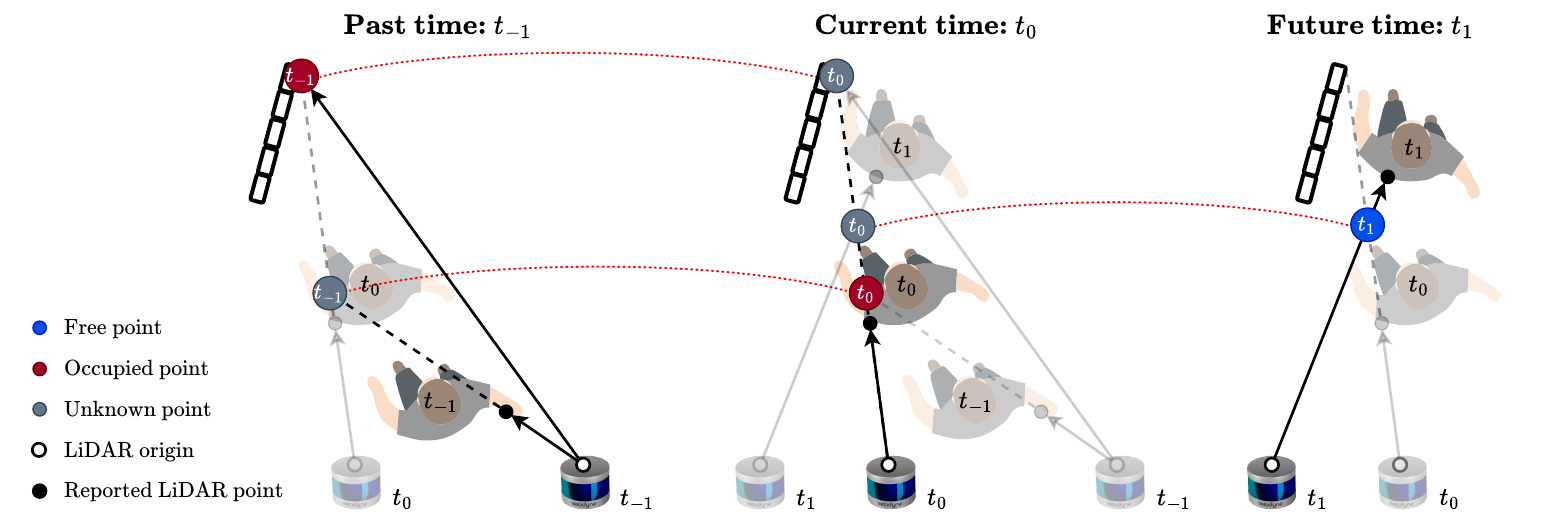}
    \caption{Temporal occupancy inconsistency caused by the relative motion between the sensor and the person. The red dotted lines track fixed points (temporal overlapping points) in space across time, illustrating how their occupancy states change. The point colors denote the occupancy state, while the index indicates the observation time.}
    \label{fig:top}
\end{figure*}

Moving Object Segmentation (MOS) on LiDAR data aims to segment moving points from the entire LiDAR scan, providing autonomous systems~\cite{lmnet,4dmos,mapmos,insmos, mdetector} with important clues for localization and dynamic obstacle avoidance, thereby enabling better planning of collision-free trajectories in dynamic scenes. However, per-point manual labeling of moving objects~\cite{lmnet,sekitti,nuscenes} is both costly and labor-intensive.
Meanwhile, self-supervised pre-training has achieved significant success in natural language processing~\cite{bert,roberta,deberta,gpt1,gpt3}, computer vision~\cite{unsupervised_videos,moco,mae,vit,clip}, and 3D vision~\cite{pointbert,pointmae,pointcontrast,unipad}, enabling the learning of transferable embeddings from unlabeled raw data. This motivates us to pre-train a backbone using raw LiDAR data in a self-supervised manner and then fine-tune it for MOS, where labeled data is limited and expensive.

The MOS task heavily relies on temporal information to discern the motion of objects relative to their surroundings. 
In the literature involving temporal information of LiDAR data, an intuitive approach is to utilize LiDAR observation forecasting as a self-supervised pre-training method for MOS.
The forecasting task includes point cloud forecasting ~\cite{s2net,spfnet,st3dcnn,copilot4d,4docc} and occupancy forecasting~\cite{uno}, which focus on predicting future scans or occupancy maps based on past and current LiDAR scans and poses.
However, utilizing forecasting as a pre-training method for MOS presents certain limitations: \textbf{(1)} The forecasting task is more complex and advanced than MOS. Forecasting a static object mainly requires geometric information to reconstruct the unseen parts of the object from a novel sensor view. For moving objects, beyond geometry, it is essential to extract temporal features such as the object's kinematics and interactions between objects and their surroundings. Hence, relying on forecasting as a pre-training step for MOS could be unnecessarily complex. \textbf{(2)} Existing methods~\cite{4docc,uno} inevitably learn a ``flow" when using raw future observations as supervision. This complex ``flow" information represents the relationship between the embeddings of past observations and future observations. Current approaches either implicitly learn this flow through simple feature broadcasting~\cite{4docc} or explicitly predict it~\cite{uno}, lacking direct supervision similar to the scene flow task~\cite{deflow,seflow,icpflow}. Thus, the learned ``flow" tends to be noisy. Relying on this noisy ``flow" for deterministic future observation predictions (as opposed to the generative approach~\cite{copilot4d}) can lead to network overfitting, thereby reducing its transferability to the downstream task.

To overcome these two limitations and incorporate temporal information for MOS self-supervised pre-training, we propose \textbf{TOP}, namely \textbf{T}emporal \textbf{O}verlapping \textbf{P}rediction. \textbf{TOP} explores the occupancy states of temporal overlapping points observed by both the current scan and adjacent (past or future) scans for pre-training. The rationale is that the motion of the sensor and objects creates temporal occupancy inconsistency, where a temporal overlapping point may exhibit different occupancy states when observed at different times, as shown in Fig~\ref{fig:top}.
We first pre-process the spatial positions and occupancy states of the temporal overlapping points by considering beam divergence~\cite{divergence}, which is a physical characteristic of LiDAR. Leveraging this physical characteristic of LiDAR helps avoid discretization errors caused by manually setting a fixed voxel resolution. Then, we pre-train the encoder by predicting the pre-processed occupancy states, which simplifies the task compared to directly forecasting LiDAR observations. Based on a distinct temporal correlation between LiDAR beams at different times established by the pre-processing, our pre-training avoids the need for either implicitly or explicitly learning a noisy ``flow". Additionally, since understanding the current scene geometry is also important for MOS, we include a reconstruction term in our pre-training objective.

We evaluate \textbf{TOP} via few-shot fine-tuning on nuScenes~\cite{nuscenes} with varying amounts of labeled data. The results demonstrate that \textbf{TOP} outperforms both the supervised training-from-scratch baseline~\cite{4dmos} and other self-supervised pre-training baselines, including state-of-the-art (SOTA) LiDAR forecasting methods~\cite{4docc,uno} and the occupancy-based 3D method~\cite{also}. Furthermore, cross-dataset experiments on SemanticKITTI~\cite{sekitti} highlight \textbf{TOP}'s superior transferability, while its few-shot performance on the nuScenes semantic segmentation task demonstrates its potential to generalize to other downstream tasks.
\section{Related Works}
\noindent\textbf{Moving Object Segmentation.} Existing supervised MOS methods can be grouped into projection-based and voxel-based approaches. The early MOS methods~\cite{lmnet,rvmos,motionseg3d,mfmos} project 3D LiDAR data onto range images and embed the range residual images using a 2D backbone. These methods suffer from imprecise boundaries when re-projecting 2D outputs to 3D point clouds~\cite{mapmos}. Recent methods~\cite{pointmoseg,4dmos,mapmos,insmos} utilize a sparse convolution backbone to directly embed voxelized point clouds. 4DMOS~\cite{4dmos} first employed sparse 4D UNet~\cite{mink} and established it as the SOTA backbone for MOS. For unsupervised approaches, several scene flow methods~\cite{egomotion,slim,weakly,ssfmos} decouple the per-point flow into an ego-motion flow for static points and a non-rigid flow for moving points, which yields MOS as a byproduct. However, these two-frame approaches encounter challenges in identifying slowly moving objects~\cite{4dmos}. MOTS~\cite{mots} proposes a label-free method that is limited to stationary scenes. 4dNDF~\cite{4dndf} encodes the 4D scene into an implicit neural map, which can be used to distinguish between the static map and dynamic objects. However, its computational cost limits real-time applicability. In addition, M-detector~\cite{mdetector} proposes a network-free approach based on the occlusion principle. It requires manual tuning of numerous hyperparameters when using different LiDARs or deploying in different scenes. 
To reduce manual labeling costs, several methods~\cite{automos,helimos} propose automatic labeling pipelines, though some of them still incorporate a manual refinement step.
In contrast to previous MOS works, we employ self-supervised pre-training to enhance the few-shot MOS performance.

\noindent \textbf{3D Self-supervised Pre-training.} Effective self-supervised pre-training methods in natural language processing~\cite{bert, gpt1, gpt3} and computer vision~\cite{mae} have been extended to 3D data. Some methods~\cite{pointbert,posbert,pointgpt} leverage BERT-like~\cite{bert} or GPT-like~\cite{gpt1,gpt3} pre-training on object CAD point clouds. Contrastive learning methods~\cite{pointcontrast, bevcontrast, strl, gcc3d, co3} employ various transformations or augmentations for contrastive losses at point, instance, and scene levels, learning geometric or semantic information without manual labels.
Inspired by MAE~\cite{mae}, numerous methods~\cite{pointmae, pointm2ae, voxelmae, gdmae, geomae, occupancymae, maeli, mvjar} pre-train an encoder by reconstructing masked point cloud patches, which yields effective 3D representations for object detection and semantic segmentation.
Subsequent works~\cite{ponder,ponderv2,unipad} apply NeRF-like~\cite{nerf} volumetric neural rendering to integrate 2D and 3D multi-modal data. To the best of our knowledge, while a few methods~\cite{tarl,tmae} incorporate temporal information, \textbf{TOP} is the first self-supervised pre-training method specifically designed for dynamic tasks like MOS.

\noindent \textbf{LiDAR Observation Forecasting.} Existing works that involve temporal information of LiDAR data mainly focus on LiDAR observation forecasting, including point cloud and occupancy forecasting. Previous works on point cloud forecasting~\cite{s2net,spfnet,st3dcnn} directly predict the future LiDAR scans based on past ones. A recent study, 4DOcc~\cite{4docc}, highlights the potential of forecasting as a scalable self-supervised task. Instead of directly predicting raw point clouds, 4DOcc renders the future point clouds on a predicted occupancy map using known sensor intrinsics and extrinsics, enabling the network to focus on learning the world rather than sensor characteristics. Another generative approach, Copilot4D~\cite{copilot4d}, introduces the concept of the world model. It tokenizes LiDAR scans and predicts future scans through discrete diffusion. However, the learned world knowledge is difficult to apply to downstream tasks due to its heavy-decoder architecture. ALSO~\cite{also} is a single-scan method that reconstructs the occupancy of a local surface. This work inspired UnO~\cite{uno}, which is the most recent occupancy forecasting method that predicts the binary occupancy class of points sampled along future LiDAR rays. In contrast, our work diverges from these forecasting-based approaches by pre-training the backbone with our proposed \textbf{TOP} method.
\section{Methodology}
\label{sec:method}

\begin{figure}[ht]
    \centering
    \includegraphics[width = 1\linewidth]{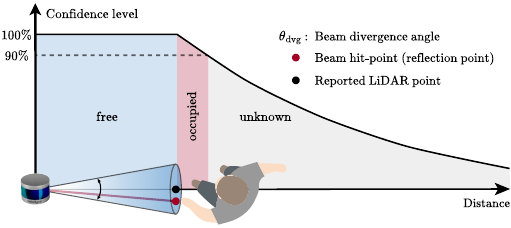}
    \caption{\textbf{LiDAR beam divergence:} The diverging beam is illustrated as the blue cone. The reported point (black) lies on the beam's centerline, while the actual beam hit-point (red) can be anywhere within the beam's footprint.
    \textbf{LiDAR occupancy measurement:} We model the space from the sensor to the reported LiDAR point as free with maximum confidence. Beyond this point, confidence decays exponentially. The region immediately following the reported point is considered occupied if its confidence is above a predefined threshold. The subsequent region where the confidence falls below the threshold is considered unknown.
    }
    \label{fig:exp_decay}
\end{figure}

\begin{figure*}[ht]
    \centering
    \includegraphics[width=1\linewidth]{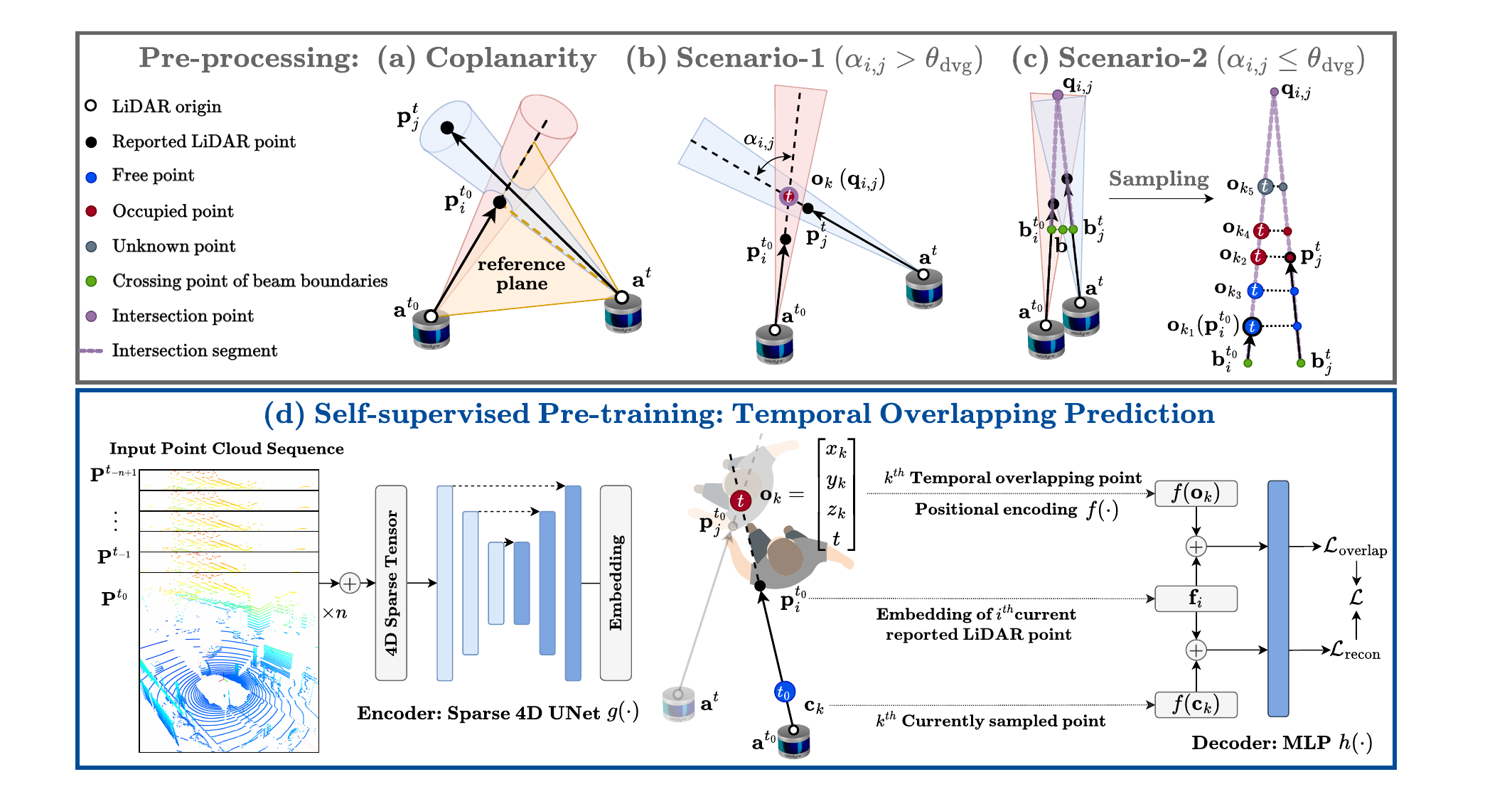}
    \caption{Overall pipeline. \textbf{Pre-processing}: \textbf{(a)} shows the coplanarity condition of two beams. \textbf{(b)} If the spatial angle between two beams $\alpha_{i,j}$ exceeds the beam divergence angle $\theta_{\text{dvg}}$, we sample the beams' intersection point as the temporal overlapping point. The red and blue areas indicate beam divergence. \textbf{(c)} When $\alpha_{i,j}$ is less than $\theta_{\text{dvg}}$, we calculate the intersection segment and sample temporal overlapping points from it. \textbf{Pre-training}: \textbf{(d)} The input sequence is encoded by a sparse 4D UNet. For both pre-training objectives, a shallow MLP decoder predicts occupancy states based on the point's positional encoding and the feature of its corresponding LiDAR point.}
    \label{fig:pipeline}
\end{figure*}

In this section, we introduce \textbf{TOP}, a self-supervised method to pre-train an encoder for the downstream MOS task. The pre-training goal involves two objectives: predicting the occupancy of temporal overlapping points and reconstructing the current scene's occupancy. We first introduce the notation and problem formulation in Sec.~\ref{subsec:formulation}. Then, we discuss the pre-processing of the temporal overlapping points in Sec.~\ref{subsec:preprocessing}. Finally, we describe the pre-training objectives in Sec.~\ref{subsec:objectives}.

\subsection{Problem Formulation}
\label{subsec:formulation}
\noindent\textbf{Notations.} To begin with, we define a LiDAR scan as $\mathbf{P} \in \mathbb{R}^{N \times 3}$, where $N$ is the number of points.  We use subscript $i$ to indicate the $i^{th}$ point in $\mathbf{P}$, where each point $\mathbf{p}_i\in \mathbb{R}^{3}$ ($i=1,2, \dots, N$) contains $xyz$ coordinates. $\mathbf{p}_i$ can be further described by a laser beam emitted from the sensor origin $\mathbf{a}\in\mathbb{R}^3$ with its normalized beam direction $\mathbf{d}_i\in\mathbb{R}^3$ and range $r_i$, leading to $\mathbf{p}_i=\mathbf{a}+r_i\mathbf{d}_i$. Although the laser beam is commonly modeled as an ideal ray, an actual LiDAR beam exhibits a diverging radius as it travels, a phenomenon known as beam divergence~\cite{divergence}. The beam divergence angle, denoted as $\theta_{\text{dvg}}$, is a constant for each type of LiDAR.

When a diverging LiDAR beam hits an obstacle and reflects, the sensor reports a point on the beam's centerline corresponding to the measured time-of-flight range. In contrast, the actual beam hit-point (reflection point) can be located anywhere within the beam's footprint, as shown in Fig.~\ref{fig:exp_decay}.
To incorporate the temporal LiDAR sequence, we indicate scans at different timestamps with superscripts. For example, the scan at the current timestamp $t_0$ is $\mathbf{P}^{t_0} \in \mathbb{R}^{N_{t_0} \times 3}$. For adjacent past/future timestamps $\mathcal{T}=\{t_{-n}, \dots, t_{-1}, t_1, \dots, t_n\}$, we have $\mathcal{P}_{\text{adj}}=\{\mathbf{P}^t \in \mathbb{R}^{N_t \times 3}\}_{t\in\mathcal{T}}$. The $j^{th}$ beam at time $t$, $\mathbf{d}_j^t$, results in the point $\mathbf{p}_j^t$, which is described by $\mathbf{p}_j^t=\mathbf{a}^t+r_j^t\mathbf{d}_j^t$.

\paragraphskip
\noindent\textbf{Point Cloud Encoding.} All scans in $\mathcal{P}_{\text{adj}}$ are transformed to the local coordinate system of $\mathbf{P}^{t_0}$ using the poses provided in the datasets. We take $n$ current and past scans as network inputs $\mathcal{P}_{\text{in}} = \{\mathbf{P}^{t_{-n+1}}, \dots, \mathbf{P}^{t_{-1}}, \mathbf{P}^{t_0}\}$. Each point in $\mathcal{P}_{\text{in}}$ is concatenated with its timestamp before being fed into the network. We apply a sparse 4D UNet encoder $g(\cdot)$ used in~\cite{mink,4dmos,mapmos,insmos} to embed the spatiotemporal information of the 4D inputs:
\begin{equation}
    \mathbf{F} = g(\mathcal{P}_\text{in}),
\end{equation}
where $\mathbf{F}\in \mathbb{R}^{N_{t_0} \times d}$ is a sparse feature for points in the current scan $\mathbf{P}^{t_0}$ with a feature dimension of $d$.

\paragraphskip
\noindent\textbf{Pre-training Loss.} We incorporate temporal overlapping points and their temporal occupancy states to pre-train the encoder $g(\cdot)$. As described in Fig.~\ref{fig:top}, temporal overlapping points are locations along the LiDAR beams from the current scan $\mathbf{P}^{t_0}$ that are also intersected by beams from adjacent past/future scans $\mathcal{P}_\text{adj}$. Firstly, we denote the temporal overlapping points as $\mathbf{O} \in \mathbb{R}^{M \times 4}$, where $M$ is the number of points and each point has $xyz$ coordinates and the timestamp $t$ of the adjacent scan that observed it. Then, we use the LiDAR occupancy measurement (introduced in Sec.~\ref{subsec:preprocessing}) to determine the occupancy state labels for $\mathbf{O}$, denoted as $\mathbf{S} \in \mathbb{R}^{M \times 3}$. The overall pre-training loss $\mathcal{L}$, further described in Sec.~\ref{subsec:objectives}, consists of temporal overlapping prediction $\mathcal{L}_{\text{overlap}}$ and current occupancy reconstruction $\mathcal{L}_{\text{recon}}$:
\begin{equation}
    \mathcal{L}=\mathcal{L}_{\text{overlap}}(\mathbf{O}, \mathbf{S}, \mathbf{F}) + \mathcal{L}_{\text{recon}}(\mathbf{P}^{t_0},\mathbf{F}).
\end{equation}

\subsection{Temporal Overlapping Pre-processing}
\label{subsec:preprocessing}
As computing temporal overlapping points during pre-training can be inefficient, we pre-process the LiDAR scans $\mathbf{P}^{t_0}$ and $\mathcal{P}_\text{adj}$ to produce temporal overlapping points $\mathbf{O}$ and their occupancy states $\mathbf{S}$. We first identify coplanar beam pairs from $\mathbf{P}^{t_0}$ and $\mathcal{P}_\text{adj}$, as coplanarity is a necessary condition for beams to overlap, as illustrated in Fig.~\ref{fig:pipeline}(a). Then, we compute the intersection of coplanar beam pairs and sample temporal overlapping points. We handle the computation and sampling in two distinct scenarios based on the spatial angle between two beams, as shown in Fig.~\ref{fig:pipeline}(b) and Fig.~\ref{fig:pipeline}(c), respectively. For clarity, we consider a beam pair: the $i^{th}$ beam at time $t_0$ and the $j^{th}$ beam at time $t$ ($\mathbf{d}_i^{t_0}$, $\mathbf{d}_j^{t}$), emitted from ($\mathbf{a}^{t_0}$, $\mathbf{a}^t$) and hitting ($\mathbf{p}_i^{t_0}$, $\mathbf{p}_j^t$). When a beam pair ($\mathbf{d}_i^{t_0}$, $\mathbf{d}_j^{t}$) generates the $k^{th}$ temporal overlapping point $\mathbf{o}_k$, we record the $(i,j,k)$ correspondence for subsequent usage in Sec.~\ref{subsec:objectives}.

\paragraphskip
\noindent\textbf{Coplanarity.} We define a reference plane formed by ($\mathbf{a}^{t_0}$, $\mathbf{a}^t$, $\mathbf{p}_i^{t_0}$), illustrated as the yellow triangle in Fig.~\ref{fig:pipeline}(a). The unit normal vector of this plane is:
\begin{equation}
\mathbf{n}_i^t=\mathbf{d}_i^{t_0} \times \frac{\mathbf{a}^t}{||\mathbf{a}^t||}.
\end{equation} The spatial angle between beam $\mathbf{d}_j^t$ and the plane is:
\begin{equation}
\mathbf{\theta}_{i,j}=\arccos{(\mathbf{n}_i^t \cdot \mathbf{d}_j^t)} - \frac{\pi}{2}.
\end{equation}
Based on the physical beam divergence, if $|\mathbf{\theta}_{i,j}| \leq \frac{1}{2}\theta_{\text{dvg}}$, we assume the two beams ($\mathbf{d}_i^{t_0}$, $\mathbf{d}_j^t$) are coplanar. All coplanar beam pairs are split into two scenarios based on the value of their spatial angle $\alpha_{i,j}=\arccos{(\mathbf{d}_i^{t_0} \cdot \mathbf{d}_j^t)}$.

\noindent\textbf{Scenario-1 (}$\alpha_{i,j} > \theta_{\text{dvg}}$\textbf{).} This scenario describes most overlapping cases where the spatial angle between two beams exceeds the beam divergence angle, as shown in Fig.~\ref{fig:pipeline}(b). The beam intersection is small and is thus considered a point $\mathbf{q}_{i,j}$, which is the intersection of the two beams' centerlines. $\mathbf{q}_{i,j}$ is calculated from the parameter equations of the two beams:
\begin{equation}
    \mathbf{q}_{i,j} = \frac{\mathbf{a}^{t} \times \mathbf{d}_j^t \cdot \mathbf{m}_{i,j}}{\mathbf{m}_{i,j} \cdot \mathbf{m}_{i,j}} \cdot \mathbf{d}_i^{t_0},
\end{equation}
where $\mathbf{m}_{i,j} = \mathbf{d}_i^{t_0} \times \mathbf{d}_j^t$ is a vector along the common perpendicular to the two beams. In this scenario, the intersection point $\mathbf{q}_{i,j}$ is selected as the $k^{th}$ temporal overlapping point $\mathbf{o}_k$.

\noindent\textbf{Scenario-2 (}$\alpha_{i,j} \leq \theta_{\text{dvg}}$\textbf{).} When the spatial angle between two beams is less than or equal to the beam divergence, the intersection of the two beams can no longer be considered a point but rather a region, as shown in Fig.~\ref{fig:pipeline}(c). In this scenario, we compute the intersection segments (purple dashed lines) on the centerlines for subsequent sampling. Firstly, we obtain the endpoint $\mathbf{q}_{i,j}$ from the beam intersection similar to Scenario-1. Then, we approximate the starting point $\mathbf{b}_i^{t_0}$ and $\mathbf{b}_j^{t}$ as the projection from a point $\mathbf{b}$ onto the two beams. Here, $\mathbf{b}$ is the crossing point of beam boundaries. Given that the spatial angle $\alpha_{i,j}$ is small, we make two approximations to compute the starting points: (1) We approximate that the two intersection segments share the same length. (2) We approximate that the distance between the two starting points $\mathbf{b}_i^{t_0}$ and $\mathbf{b}_j^{t}$ is equal to the sum of the beam radii at the two points. The detailed formulas for computing $\mathbf{b}_i^{t_0}$ and $\mathbf{b}_j^{t}$ are provided in the appendix.

Since the overlap between two beams is a region, we sample 5 temporal overlapping points within the intersection segment, as shown in Fig.~\ref{fig:pipeline}(c)-right: $\mathbf{o}_{k_1}$, the beam hit point $\mathbf{p}_i^{t_0}$; $\mathbf{o}_{k_2}$, the projection of $\mathbf{p}_j^{t}$ on $\mathbf{d}_i^{t_0}$; $\mathbf{o}_{k_3}$, the midpoint of ($\mathbf{o}_{k_1}$, $\mathbf{o}_{k_2}$); $\mathbf{o}_{k_4}$, the midpoint of ($\mathbf{o}_{k_1}$, $\mathbf{q}_{i,j}$); $\mathbf{o}_{k_5}$, the midpoint of ($\mathbf{o}_{k_2}$, $\mathbf{q}_{i,j}$). The occupancy state for each of these sampled points at timestamp $t$ is determined by projecting it onto the adjacent beam $\mathbf{d}_j^{t}$ and evaluating the corresponding occupancy.

\paragraphskip
\noindent\textbf{LiDAR Occupancy Measurement.} As shown in Fig.~\ref{fig:exp_decay}, the LiDAR occupancy measurement~\cite{uno, analytical} partitions the space along each beam into three states. First, the space from the sensor origin to the reported point is considered free with maximum confidence. Beyond the reported point, an exponential confidence decay defines two subsequent states: the space is occupied while the confidence exceeds a predefined threshold, and is considered unknown where the confidence drops below the threshold.

Given the $k^{th}$ temporal overlapping point $\mathbf{o}_k$ generated from the overlapping beam pair ($\mathbf{d}_i^{t_0}$, $\mathbf{d}_j^t$), the confidence level $w_{\text{conf}}$ of $\mathbf{o}_k$ is formulated as:
\begin{equation}
    w_{\text{conf}}(\mathbf{o}_k,\mathbf{p}_j^t) = \begin{cases}
        1, &\text{if}\ ||\mathbf{o}_k||\leq||\mathbf{p}_j^t||, \\
        e^{{-(||\mathbf{o}_k|| - ||\mathbf{p}_j^t||)}}, &\text{if}\ ||\mathbf{o}_k||>||\mathbf{p}_j^t||. \\
    \end{cases}
    \label{eq:conf}
\end{equation}
The occupancy state of $\mathbf{o}_k$ at time $t$ is denoted as $\mathbf{s}_k$:
\begin{equation}
    \mathbf{s}_k = \begin{cases}
        [1, 0, 0], &\text{if}\ ||\mathbf{o}_k||<||\mathbf{p}_j^t||, \\
        [0, 1, 0], &\text{if}\ ||\mathbf{o}_k||=||\mathbf{p}_j^t||\ \text{or}\ \lambda_{\text{occ}} \leq w_{\text{conf}}(\mathbf{o}_k,\mathbf{p}_j^t) < 1, \\
        [0, 0, 1], &\text{if}\ w_{\text{conf}}(\mathbf{o}_k,\mathbf{p}_j^t)< \lambda_{\text{occ}}, \\
    \end{cases}
    \label{eq:state}
\end{equation}
where $\mathbf{s}_k$ is a one-hot vector indicating one of three occupancy states: free, occupied, or unknown. $\lambda_{\text{occ}}$ is the predefined confidence threshold for the occupied state.

\subsection{Training Objectives}
\label{subsec:objectives}
Our primary pre-training objective, temporal overlapping prediction, is to predict the occupancy state $\mathbf{s}_k$ of $k^{th}$ temporal overlapping point $\mathbf{o}_k$. We predict the occupancy based on the positional encoding of $\mathbf{o}_k$ and the $i^{th}$ current LiDAR point's feature $\mathbf{f}_i$ by a shallow MLP decoder $h(\cdot)$:
\begin{equation}
    \mathbf{u}_k = h(f(\mathbf{o}_k) + \mathbf{f}_i) \in \mathbb{R}^{3},
    \label{eq:decoder}
\end{equation}
where $\mathbf{u}_k$ is the predicted probability over the three occupancy states, and $f(\mathbf{o}_k) \in \mathbb{R}^{d}$ denotes the sine-cosine 4D positional encoding vector as described in~\cite{transformer}. Temporal overlapping prediction is formulated as a Cross-Entropy loss:
\begin{equation}
    \mathcal{L}_{\text{overlap}} = -\frac{1}{M}\sum\limits_{k=1}^{M} w_{\text{conf}}(\mathbf{o}_k,\mathbf{p}_j^t) \cdot \mathbf{w}_s \cdot \mathbf{s}_{k} \cdot \log{(\mathbf{u}_{k})},
\end{equation}
where $\mathbf{w}_s$ is a constant weight vector for each occupancy state, and the $(i,j,k)$ correspondence was introduced in Sec.~\ref{subsec:preprocessing}.

Additionally, we reconstruct the current occupancy using the input scans, which serves as a regularization term in the overall loss. We randomly sample $M'$ occupied and free points along each current beam. For the $k^{th}$ point $\mathbf{c}_k$ sampled on the $i^{th}$ current beam, it consists of a spatial position and timestamp $t_0$. The ground truth occupancy label is $\mathbf{s}_k$, and the predicted probability is $\mathbf{v}_k$. The reconstruction objective is also formulated as a Cross-Entropy loss:
\begin{align}
    \mathbf{v}_k &= h(f(\mathbf{c}_k) + \mathbf{f}_i) \in \mathbb{R}^{3}, \\
    \mathcal{L}_{\text{recon}} &= -\frac{1}{N_{t_0}M'} \sum\limits_{i=1}^{N_{t_0}} \sum\limits_{k=1}^{M'} \mathbf{w}_{s} \cdot \mathbf{s}_k \cdot  \log{(\mathbf{v}_{k})}.
\end{align}
\section{Evaluation Metrics}
\label{sec:metrics}
We introduce a robust evaluation for MOS, designed to overcome two critical biases in the conventional IoU metric~\cite{lmnet,4dmos,mapmos,insmos}: the over-weighting of objects with more points, and the inflation caused by the ego vehicle's points.

A primary issue with the conventional IoU is its inherent bias towards objects with more points, leading to the neglect of smaller or distant objects. This poses significant safety risks, as all dynamic objects are equally critical for safe operation, regardless of their point count. For instance, a high-speed vehicle might fail to detect a distant car, or a UAV could miss a fast-approaching object. To quantify this imbalance, our statistical analysis of the nuScenes dataset reveals that the smallest 75.22\% of objects (by point count) account for only 15.49\% of the total points (see appendix). This confirms that a few large objects heavily dominate the final IoU score. The second bias is the inflation of IoU. In practice, the ego vehicle's state is known via onboard sensors, and is not part of the perception challenge. Including these easy-to-predict points (which account for 76.72\% of all moving points in the nuScenes dataset) severely inflates the metric, masking the model's actual performance on external dynamic objects.

To address these limitations, we use two complementary metrics. First, to counteract the object-size bias, we use an object-level metric $\text{Recall}_{\text{obj}}$, which ensures an equal weight of each moving object by averaging their individual recalls. It is computed as:
\begin{equation}
    \text{Recall}_{\text{obj}} = \frac{1}{m} \sum\limits_{i=1}^{m} \frac{\text{TP}_i}{\text{TP}_i + \text{FN}_i},
\end{equation}
where $m$ is the number of moving objects, while $\text{TP}_i$ and $\text{FN}_i$ are the number of true positive and false negative points, respectively, within the ground truth bounding box of the $i^{th}$ object. Second, to resolve the ego-vehicle bias, we simply exclude all ego points from the IoU calculation and rename the metric $\text{IoU}_{\text{w/o}}$. This provides a more focused assessment of the model's performance on external dynamic objects. Notably, for the SemanticKITTI benchmark, $\text{IoU}_{\text{w/o}}$ is equivalent to the conventional IoU, as the benchmark already omits ego points by classifying them as unlabeled. To demonstrate that the conventional IoU metric fails to effectively evaluate perception of external dynamic objects when numerous ego-vehicle points are included, we provide supplementary results on nuScenes using conventional IoU in the appendix.

In summary, our dual-metric evaluation provides a holistic and unbiased assessment. The two metrics are complementary: $\text{Recall}_{\text{obj}}$ ensures that the detection completeness of each object is fairly evaluated, preventing smaller ones from being overlooked, while $\text{IoU}_{\text{w/o}}$ penalizes false positive predictions in the surrounding scene.
\section{Experiments}
\label{sec:exp}
\subsection{Experiments Setup}
\noindent\textbf{Datasets.} We pretrain the encoder on the nuScenes dataset~\cite{nuscenes}, which utilizes a 32-beam Velodyne LiDAR for data collection. The beam divergence angle $\theta_{\text{dvg}}$ specified by the manufacturer is 0.003 rad. We perform few-shot experiments on the nuScenes dataset. Previous methods~\cite{mapmos,pointmoseg} directly convert nuScenes object attributes into MOS labels. However, these attributes do not precisely describe the object's motion state. Moreover, some objects lack attributes and are thus overlooked during training or evaluation. To overcome this limitation, we refine MOS labels by incorporating object attributes and speed. We compute object speed from bounding box annotations and use it to classify motion states based on predefined static and moving thresholds. We introduce a speed range between these thresholds to categorize objects with ambiguous motion states into an unknown class, such as cars that are just starting or hesitant pedestrians. More details are available in the appendix. We also conduct few-shot cross-dataset transfer experiments on the SemanticKITTI~\cite{sekitti} MOS benchmark~\cite{lmnet}, which employs a 64-beam Velodyne LiDAR for data collection.

\noindent\textbf{Implementation Details.}
In our setup, we set $n=6$, where $\mathcal{P}_{\text{adj}}$ includes 6 past scans and 6 future scans, and the input sequence $\mathcal{P}_{\text{in}}$ comprises 6 scans. All scans are sampled at 2Hz. The confidence cutoff threshold for the occupied state is $\lambda_{\text{occ}}=0.9$. We apply data augmentation techniques from~\cite{segcontrast,4dmos} and discard points outside a bounded area based on~\cite{4docc,uno}. Our encoder is a sparse 4D UNet. Following previous works~\cite{4dmos,mapmos}, we initialize all point features to a constant. The quantization size for the sparse backbone is 0.1~m, and the feature dimension $d=128$. For our temporal overlapping prediction objective, we address class imbalance due to the scarcity of occupied samples. Specifically, in each training iteration, we use all available occupied samples, while randomly sampling five times as many points for the free class and the same number of points for the unknown class. As for the current occupancy reconstruction objective, we sample 5 occupied points and 25 free points on each current beam ($M'=30$). The class weight is $\mathbf{w}_s=[1,5,1]$ for free, occupied, and unknown classes. We use a fixed random seed for all experiments to ensure reproducibility. Detailed optimizer settings are in the appendix.

\noindent\textbf{Baselines.} We benchmark our method against two classes of baselines: (1) a training-from-scratch baseline using the sparse 4D UNet from~\cite{4dmos,mapmos,insmos}, and (2) self-supervised pre-training baselines, including the LiDAR forecasting methods 4DOcc~\cite{4docc} and UnO~\cite{uno}, and the occupancy-based 3D pre-training method ALSO~\cite{also}. To ensure a fair comparison, we replace their native encoders with the sparse 4D UNet while retaining their original decoders.

\subsection{Few-shot Results on nuScenes}
\label{subsec:mos_exp}
We evaluate few-shot MOS on nuScenes. We first pre-train an encoder on the entire training split. Then, the encoder is fine-tuned for point-wise MOS prediction using a binary Cross-Entropy loss on randomly sampled subsets (5\%, 10\%, 20\%, 50\%) of the nuScenes training sequences. All models are trained to convergence. For evaluation, we select the best-performing checkpoint for each of our two metrics separately. This model selection is performed on a validation set created by randomly partitioning the official nuScenes training split. The final results are reported on the official nuScenes validation split (used as our test set) in Tab.~\ref{tab:nusc-mos}.

\textbf{TOP} outperforms the training-from-scratch baseline across all data subsets. It demonstrates substantial gains in $\text{Recall}_{\text{obj}}$, achieving up to 12.17\% relative improvement (from 24.98\% to 28.03\%) on the 10\% data subset. While the gains in $\text{IoU}_{\text{w/o}}$ are more modest, we attribute this to \textbf{TOP}'s enhanced ability to detect small moving objects, such as pedestrians and cyclists. These objects significantly contribute to recall but constitute a small fraction of the total points, thus having a lesser impact on IoU.

\begin{table}[ht]
\centering
\setlength\tabcolsep{4pt}
\scalebox{0.8}{
\begin{tabular}{c|c|cc|cc}
\toprule
\multirow{2}*{Data} & \multirow{2}*{Pretrain} & \multicolumn{2}{c|}{Best $\textbf{Recall}_{\textbf{obj}}$} & \multicolumn{2}{c}{Best $\textbf{IoU}_{\textbf{w/o}}$} \\
\cmidrule(lr){3-4} \cmidrule(lr){5-6}
& & $\textbf{Recall}_{\textbf{obj}}$ & $\text{IoU}_{\text{w/o}}$ & $\text{Recall}_{\text{obj}}$ & $\textbf{IoU}_{\textbf{w/o}}$ \\
\midrule
\multirow{5}{*}{5\%}         & No & 24.27 & 34.96 & 23.20 & 35.05      \\
                             & ALSO                                    & $\text{18.68}^{\text{\red{-5.59}}}$       & $\text{31.65}^{\text{\red{-3.31}}}$       & $\text{17.57}^{\text{\red{-5.63}}}$       & $\text{31.97}^{\text{\red{-3.07}}}$       \\
                             & 4DOcc                                   & $\text{22.14}^{\text{\red{-2.13}}}$       & $\text{33.39}^{\text{\red{-1.57}}}$       & $\text{21.02}^{\text{\red{-2.17}}}$       & $\text{33.81}^{\text{\red{-1.24}}}$       \\
                             & UnO                                     & $\text{25.23}^{\text{\green{+0.96}}}$     & $\text{34.31}^{\text{\red{-0.65}}}$       & $\text{22.62}^{\text{\red{-0.58}}}$     & $\text{34.90}^{\text{\red{-0.15}}}$     \\
                             & \textbf{TOP}                           & $\textbf{25.75}^{\textbf{\green{+1.48}}}$ & $\textbf{35.94}^{\textbf{\green{+0.98}}}$ & $\textbf{25.75}^{\textbf{\green{+2.56}}}$ & $\textbf{35.94}^{\textbf{\green{+0.90}}}$ \\
\midrule
\multirow{5}{*}{10\%}
                             & No & 24.98 & 36.44 & 22.98 & 38.06      \\
                             & ALSO                                    & $\text{27.32}^{\text{\green{+2.34}}}$       & $\text{35.74}^{\text{\red{-0.70}}}$       & $\text{22.47}^{\text{\red{-0.51}}}$       & $\textbf{38.33}^{\textbf{\green{+0.27}}}$       \\
                             & 4DOcc                                   & 
                             $\text{24.31}^{\text{\red{-0.67}}}$       & $\text{36.01}^{\text{\red{-0.43}}}$     & $\text{23.83}^{\text{\red{+0.86}}}$       & $\text{37.46}^{\text{\red{-0.70}}}$       \\
                             & UnO                                     & $\text{26.38}^{\text{\green{+1.40}}}$     & $\text{36.56}^{\text{\green{+0.12}}}$       & $\text{22.00}^{\text{\red{-0.98}}}$       & $\text{37.17}^{\text{\red{-0.89}}}$       \\
                             & \textbf{TOP}                           & $\textbf{28.03}^{\textbf{\green{+3.04}}}$ & $\textbf{36.95}^{\textbf{\green{+0.51}}}$ & $\textbf{24.69}^{\textbf{\green{+1.72}}}$ & $\text{38.06}^{\text{\green{+0.00}}}$ \\
\midrule
\multirow{5}{*}{20\%}        & No & 25.59 & 44.25 & 24.52 & \textbf{45.05}      \\
                             & ALSO                                    & $\text{26.68}^{\text{\green{+1.08}}}$       & $\text{40.84}^{\text{\red{-3.41}}}$       & $\text{25.81}^{\text{\green{+1.28}}}$       & $\text{42.33}^{\text{\red{-2.72}}}$       \\
                             & 4DOcc                                   & $\text{26.01}^{\text{\green{+0.42}}}$     & $\text{42.21}^{\text{\red{-2.04}}}$       & $\text{24.58}^{\text{\green{+0.06}}}$     & $\text{43.11}^{\text{\red{-1.94}}}$       \\
                             & UnO                                     & $\text{28.14}^{\text{\green{+2.55}}}$     & $\text{43.63}^{\text{\red{-0.62}}}$       & $\text{26.78}^{\text{\green{+2.26}}}$     & $\text{43.97}^{\text{\red{-1.08}}}$       \\
                             & \textbf{TOP}                           & $\textbf{28.45}^{\textbf{\green{+2.86}}}$ & $\textbf{44.30}^{\textbf{\green{+0.05}}}$ & $\textbf{27.29}^{\textbf{\green{+2.76}}}$ & $\text{44.52}^{\text{\red{-0.53}}}$              \\
\midrule
\multirow{5}{*}{50\%}        & No & 29.80 & 49.29 & 29.80 & 49.29      \\
                             & ALSO                                    & $\text{30.72}^{\text{\green{+0.91}}}$       & $\text{47.34}^{\text{\red{-1.95}}}$       & $\text{28.06}^{\text{\red{-1.75}}}$       & $\text{48.10}^{\text{\red{-1.19}}}$       \\
                             & 4DOcc                                   & 
                             $\text{29.20}^{\text{\red{-0.60}}}$       & $\text{46.73}^{\text{\red{-2.56}}}$       & $\text{28.19}^{\text{\red{-1.62}}}$       & $\text{48.19}^{\text{\red{-1.10}}}$       \\
                             & UnO                                     & $\text{29.54}^{\text{\red{-0.26}}}$       & $\text{48.59}^{\text{\red{-0.70}}}$       & $\text{28.81}^{\text{\red{-1.00}}}$       & $\text{48.76}^{\text{\red{-0.53}}}$       \\
                             & \textbf{TOP}                           & $\textbf{30.58}^{\textbf{\green{+0.78}}}$ & $\textbf{49.80}^{\textbf{\green{+0.51}}}$ & $\textbf{30.58}^{\textbf{\green{+0.78}}}$ & $\textbf{49.80}^{\textbf{\green{+0.51}}}$ \\
\bottomrule
\end{tabular}
}
\caption{Few-shot MOS results on nuScenes. The best performance in each column is in bold. The metric used for model selection is in bold in the table header.}
\label{tab:nusc-mos}
\end{table}

\newpage
\subsection{Few-shot Transfer Results on SemanticKITTI}
\begin{figure}[ht]
    \centering
    \includegraphics[width = 1\linewidth]{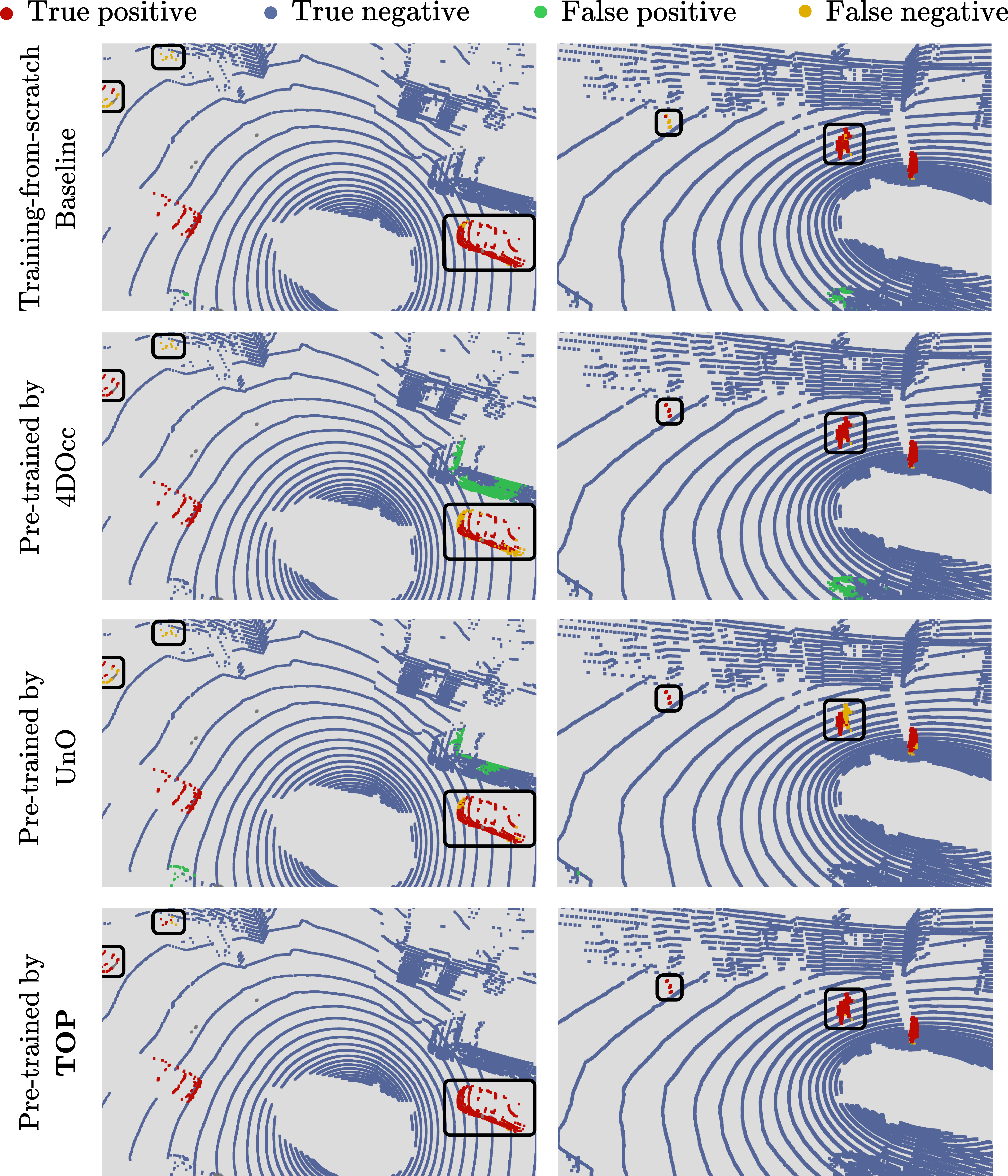}
    \caption{Qualitative results of the nuScenes MOS.}
    \label{fig:nusc-vis}
\end{figure}

\begin{figure}[ht]
    \centering
    \includegraphics[width = 1\linewidth]{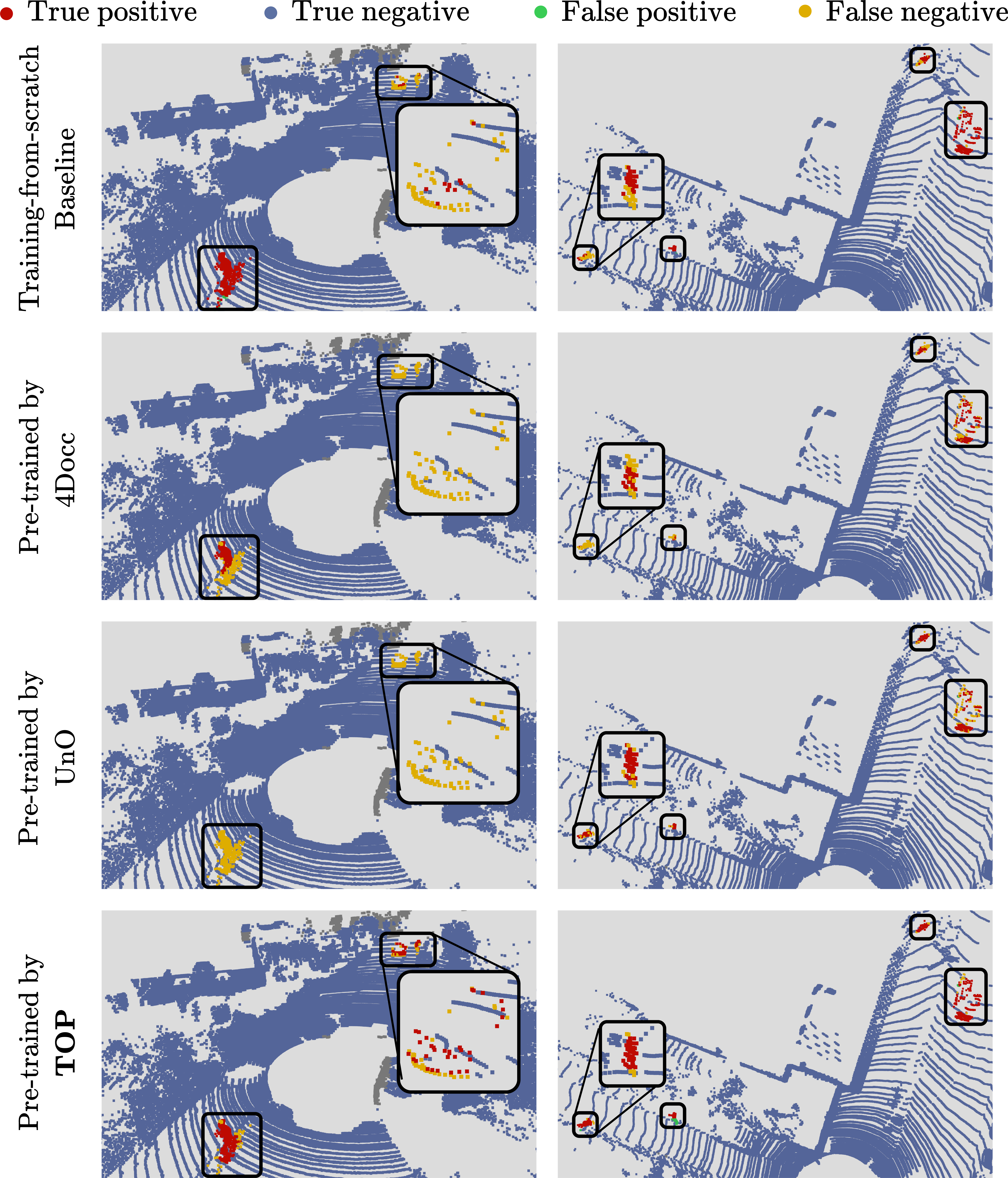}
    \caption{Qualitative results of the SemanticKITTI MOS.}
    \label{fig:kitti-vis}
\end{figure}

We evaluate the transferability of our learned spatiotemporal representations by fine-tuning the pre-trained encoder on SemanticKITTI in a few-shot setting. Specifically, we use random subsets of 10\%, 20\%, and 50\% of the official training set to fine-tune the model until convergence. We report the final results on the official SemanticKITTI validation set. As shown in Tab.~\ref{tab:kitti-mos}, \textbf{TOP} demonstrates superior transferability, significantly outperforming the training-from-scratch baseline and other pre-training baselines across all data subsets. We present qualitative MOS results on nuScenes and SemanticKITTI in Fig.~\ref{fig:nusc-vis} and Fig.~\ref{fig:kitti-vis}, respectively.

\begin{table}[ht]
\centering
\setlength\tabcolsep{4pt}
\scalebox{0.8}{
\begin{tabular}{c|c|cc|cc}
\toprule
\multirow{2}*{Data} & \multirow{2}*{Pretrain} & \multicolumn{2}{c|}{Best $\textbf{Recall}_{\textbf{obj}}$} & \multicolumn{2}{c}{Best $\textbf{IoU}_{\textbf{w/o}}$} \\
\cmidrule(lr){3-4} \cmidrule(lr){5-6}
& & $\textbf{Recall}_{\textbf{obj}}$ & $\text{IoU}_{\text{w/o}}$ & $\text{Recall}_{\text{obj}}$ & $\textbf{IoU}_{\textbf{w/o}}$ \\
\midrule
\multirow{5}{*}{10\%}
                             & No  & 37.22 & 41.22 & 33.97 & 42.81     \\
                             & ALSO                                    & $\text{9.21}^{\text{\red{-28.01}}}$       & $\text{7.33}^{\text{\red{-33.88}}}$       & $\text{9.21}^{\text{\red{-24.77}}}$       & $\text{7.76}^{\text{\red{-35.05}}}$       \\
                             & 4DOcc                                   & $\text{8.54}^{\text{\red{{-28.68}}}}$     & $\text{13.35}^{\text{\red{{-27.87}}}}$    & $\text{8.15}^{\text{\red{{-25.82}}}}$     & $\text{13.45}^{\text{\red{-29.36}}}$      \\
                             & UnO                                     & $\text{36.66}^{\text{\red{{-0.57}}}}$     & $\text{35.03}^{\text{\red{{-6.19}}}}$     & $\text{27.83}^{\text{\red{{-6.15}}}}$     & $\text{37.41}^{\text{\red{-5.41}}}$       \\
                             & \textbf{TOP}                           & $\textbf{44.68}^{\textbf{\green{+7.46}}}$ & $\textbf{43.29}^{\textbf{\green{+2.07}}}$ & $\textbf{43.36}^{\textbf{\green{+9.39}}}$ & $\textbf{49.37}^{\textbf{\green{+6.55}}}$ \\
\midrule
\multirow{5}{*}{20\%}        & No & 47.41 & 47.38 & 45.39 & 48.77      \\
                             & ALSO                                    & $\text{23.37}^{\text{\red{-24.04}}}$      & $\text{23.02}^{\text{\red{-24.36}}}$      & $\text{23.37}^{\text{\red{-22.02}}}$      & $\text{23.02}^{\text{\red{-25.75}}}$      \\
                             & 4DOcc                                   & $\text{25.61}^{\text{\red{{-21.80}}}}$    & $\text{35.83}^{\text{\red{{-11.55}}}}$    & $\text{25.23}^{\text{\red{{-20.16}}}}$    & $\text{36.24}^{\text{\red{-12.53}}}$      \\
                             & UnO                                     & $\text{45.48}^{\text{\red{-1.92}}}$       & $\text{42.09}^{\text{\red{-5.29}}}$       & $\text{43.77}^{\text{\red{-1.62}}}$       & $\text{46.09}^{\text{\red{-2.68}}}$       \\
                             & \textbf{TOP}                           & $\textbf{58.45}^{\textbf{\green{+11.04}}}$& $\textbf{50.21}^{\textbf{\green{+2.82}}}$ & $\textbf{58.45}^{\textbf{\green{+13.06}}}$& $\textbf{50.21}^{\textbf{\green{+1.44}}}$ \\
\midrule
\multirow{5}{*}{50\%}        & No & 53.13 & 56.35 & 48.20 & 59.20      \\
                             & ALSO                                    & $\text{49.82}^{\text{\red{-3.32}}}$       & $\text{57.07}^{\text{\green{+0.72}}}$     & $\text{44.27}^{\text{\red{-3.93}}}$       & $\text{60.25}^{\text{\green{+1.04}}}$   \\
                             & 4DOcc                                   & $\text{39.23}^{\text{\red{-13.91}}}$      & $\text{44.48}^{\text{\red{-11.88}}}$      & $\text{38.36}^{\text{\red{-9.84}}}$       & $\text{44.87}^{\text{\red{-14.33}}}$      \\
                             & UnO                                     & $\text{53.61}^{\text{\green{+0.48}}}$     & $\text{54.97}^{\text{\red{-1.38}}}$       & $\textbf{52.73}^{\textbf{\green{+4.53}}}$     & $\text{57.06}^{\text{\red{-2.14}}}$       \\
                             & \textbf{TOP}                           & $\textbf{55.81}^{\textbf{\green{+2.67}}}$ & $\textbf{58.85}^{\textbf{\green{+2.50}}}$ & $\text{52.11}^{\text{\green{+3.91}}}$ & $\textbf{61.55}^{\textbf{\green{+2.35}}}$     \\
\bottomrule
\end{tabular}
}
\caption{Cross-dataset few-shot transfer results on SemanticKITTI. The best performance in each column is in bold. The metric used for model selection is in bold in the table header.}
\label{tab:kitti-mos}
\end{table}

\subsection{Semantic Segmentation Results on nuScenes}
\label{subsec:semantic_exp}
We evaluate the generalization of our pre-training on a few-shot semantic segmentation task on nuScenes, by fine-tuning the pre-trained encoder using 10\% of the training split. As shown in Tab.~\ref{tab:semantic}, \textbf{TOP} surpasses the training-from-scratch baseline by 0.42\% mIoU at convergence. In contrast, models pre-trained with 4DOcc and UnO exhibit a notable performance drop. We attribute this drop to their overfitting to the complex ``flow" information during their pre-training (Sec.~\ref{sec:intro}). This result demonstrates the generalization potential of \textbf{TOP} pre-training across different downstream tasks.

\begin{table}[ht]
\centering
\setlength\tabcolsep{4pt}
\scalebox{0.8}{
\begin{tabular}{c|ccccc}
\toprule
Pretrain & No & ALSO & 4DOcc & UnO & \textbf{TOP} \\
\midrule
mIoU     & 50.55 & $50.26^{\text{\red{-0.29}}}$ & $47.49^{\text{\red{-3.06}}}$ & 
                   $48.06^{\text{\red{-2.49}}}$           & 
                   $\textbf{50.97}^{\textbf{\green{+0.42}}}$ \\
\bottomrule
\end{tabular}
}
\caption{Few-shot results on nuScenes semantic segmentation.}
\label{tab:semantic}
\end{table}

\newpage
\subsection{Ablation Study on Training Objectives}
\label{subsec:ablation}
We perform an ablation study on the two training objectives $\mathcal{L}_{\text{overlap}}$ and $\mathcal{L}_{\text{recon}}$. Results in Tab.~\ref{tab:loss} indicate that employing both objectives enhances overall performance compared to using only the temporal overlapping prediction objective. As current occupancy reconstruction is straightforward and its loss converges rapidly, we speculate it acts as a regularization term to prevent overfitting.
\begin{table}[ht]
\centering
\setlength\tabcolsep{4pt}
\scalebox{0.8}{
\begin{tabular}{c|cc|cc}
\toprule
\multirow{2}*{Objectives} & \multicolumn{2}{c|}{Best $\textbf{Recall}_{\textbf{obj}}$} & \multicolumn{2}{c}{Best $\textbf{IoU}_{\textbf{w/o}}$} \\
\cmidrule(lr){2-3} \cmidrule(lr){4-5}
& $\textbf{Recall}_{\textbf{obj}}$ & $\text{IoU}_{\text{w/o}}$ & $\text{Recall}_{\text{obj}}$ & $\textbf{IoU}_{\textbf{w/o}}$ \\
\midrule
$\mathcal{L}_\text{overlap}$ & 
24.56            &      34.79            &     22.64            &     35.35 \\
$\mathcal{L}_\text{overlap} + \mathcal{L}_\text{recon}$ &
$\textbf{25.75}$ &      $\textbf{35.94}$ &     $\textbf{25.75}$ &     $\textbf{35.94}$ \\
\bottomrule
\end{tabular}
}
\caption{Ablation study on training objectives.}
\label{tab:loss}
\end{table}
\section{Conclusion}
In this paper, we propose Temporal Overlapping Prediction (\textbf{TOP}), a self-supervised pre-training method for MOS. By learning to predict the occupancy of temporal overlapping points, \textbf{TOP} develops powerful spatiotemporal representations for moving object segmentation. We evaluate its effectiveness through few-shot experiments on nuScenes and cross-dataset transfer experiments on SemanticKITTI. \textbf{TOP} consistently improves both object-level and point-level metrics, demonstrating strong transferability across LiDAR setups. Additionally, the improved performance in semantic segmentation suggests its potential to generalize to other downstream tasks. We believe investigating \textbf{TOP} for other temporal tasks, such as object tracking, is a promising avenue for future work.

\section*{Acknowledgement}
The computations were performed using research computing facilities offered by Information Technology Services, the University of Hong Kong.

\newpage
{
    \small
    \bibliographystyle{ieeenat_fullname}
    \bibliography{main}
}

\newpage







\setcounter{section}{0}
\maketitlesupplementary

In this appendix, we describe a complementary formulation, the MOS labeling process, the cumulative distribution of moving objects and points, implementation details, the impact of ego-vehicle points on IoU, and potential negative social impacts. Finally, we present visual examples of the MOS results on the nuScenes and SemanticKITTI datasets in Fig.~\ref{fig:nuscenes} and Fig.~\ref{fig:kitti}, respectively.

\section{Formulation}
\label{sec:formulations}
As described in Sec.~3.2 Scenario-2 in the main paper, we provide the formula for the starting point $\mathbf{b}_i^{t_0}$ of the intersection segment. Since the spatial angle $\alpha_{i,j}$ is sufficiently small, we approximate that: 
(1) The two intersection segments are assumed to be equal in length, which is described as:
\begin{equation}
    ||\mathbf{q}_{i,j}|| - ||\mathbf{b}_i^{t_0}|| = ||\mathbf{q}_{i,j}-\mathbf{a}^{t}|| - ||\mathbf{b}_{j}^t-\mathbf{a}^{t}||.
    \label{eq:apprx1}
\end{equation}
(2) The distance between the two starting points $\mathbf{b}_i^{t_0}$ and $\mathbf{b}_j^{t}$ is equal to the sum of beam radii at the two points, represented as follows:
\begin{equation}
    (||\mathbf{q}_{i,j}|| - ||\mathbf{b}_i^{t_0}||) \sin{\frac{\alpha_{i,j}}{2}} = ||\mathbf{b}_i^{t_0}|| \tan{\frac{\theta_{\text{dvg}}}{2}} + ||\mathbf{b}_j^{t}-\mathbf{a}^{t}|| \tan{\frac{\theta_{\text{dvg}}}{2}}.
\label{eq:apprx2}
\end{equation}
We substitute Eq.~\ref{eq:apprx1} into Eq.~\ref{eq:apprx2} and solve for $||\mathbf{b}_i^{t_0}||$:
\begin{equation}
    ||\mathbf{b}_i^{t_0}|| = \frac{||\mathbf{q}_{i,j}|| (\sin\frac{\alpha_{i,j}}{2} + \tan{\frac{\theta_{\text{dvg}}}{2}}) - ||\mathbf{q}_{i,j}-\mathbf{a}^{t}|| \tan{\frac{\theta_{\text{dvg}}}{2}}}{\sin\frac{\alpha_{i,j}}{2} + 2 \tan{\frac{\theta_{\text{dvg}}}{2}}}.
\end{equation}

\section{MOS Labeling}
\label{sec:labeling}
As discussed in Sec.~5.1 of the main paper, the nuScenes object attributes do not precisely describe the motion state, for example, ``cycle.with\_rider" can be either moving or static. Furthermore, a proportion of objects lack attributes: $0.466\%$ of vehicles, $9.243\%$ of cycles, and $1.913\%$ of pedestrians. For our nuScenes MOS labeling, we calculate object speeds from their bounding box annotations. An object is classified as static if its speed is less than $\mu_{\text{sta}}$, and as moving if its speed exceeds $\mu_{\text{mov}}$. Objects with speeds between these thresholds are considered to have an unclear motion state and are thus classified as unknown. We use different speed thresholds for humans $(\mu_{\text{sta}}^{\text{hum}}=0.375, \mu_{\text{mov}}^{\text{hum}}=0.6)$, cycles $(\mu_{\text{sta}}^{\text{cyc}}=0.375, \mu_{\text{mov}}^{\text{cyc}}=1.0)$, and vehicles $(\mu_{\text{sta}}^{\text{veh}}=0.5, \mu_{\text{mov}}^{\text{veh}}=1.0)$.

\section{Cumulative Distribution}
\label{sec:statistics}
\begin{figure}[ht]
    \centering
    \includegraphics[width = 1\linewidth]{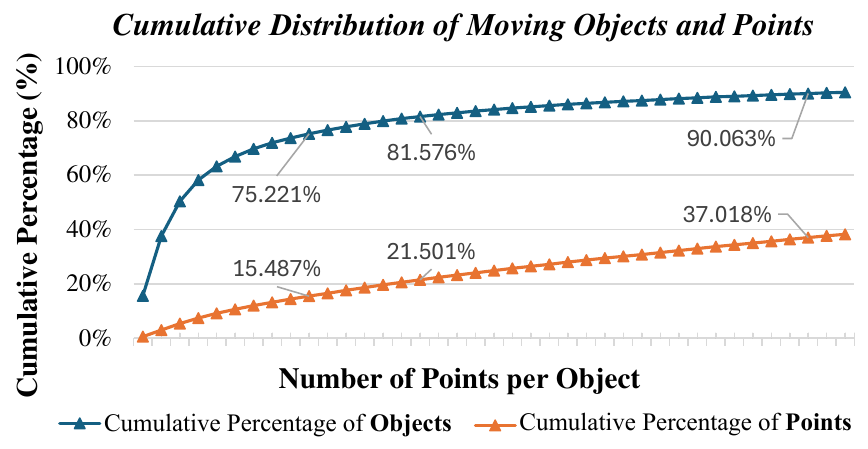}
    \caption{Cumulative distribution of moving objects and points.}
    \label{fig:cumulative}
\end{figure}

Fig.~\ref{fig:cumulative} presents a statistical analysis of moving objects from the nuScenes train-val split (discussed in Sec.~4), plotting the cumulative distribution functions of moving object instances and the number of their scanned points. The distributions are analyzed with respect to object size, defined as the number of points per object (x-axis). The significant gap between the two curves highlights a strong imbalance: the vast majority of moving objects consist of very few points. This is evidenced by the data: while moving objects with 19 or fewer points comprise 75.22\% of all moving objects, they contribute a mere 15.49\% of the total moving points. This disparity grows, with the smallest 90.06\% of objects collectively accounting for only 37.02\% of the total points. Such a skewed distribution implies that a small number of point-rich objects can disproportionately dominate the conventional IoU metric.

\section{Implementation Details}
\label{sec:details}
\begin{table}[ht]
    \centering
    \setlength\tabcolsep{4pt}
    \scalebox{0.8}{
    \begin{tabular}{c|ccccc}
    \toprule
    method & epochs & batch size & optimizer & lr start & lr max \\
    \midrule
    \textbf{TOP} & 50  & 8  & AdamW & $5.0 \times 10^{-6}$ & $5.0 \times 10^{-5}$ \\
    ALSO         & 200 & 4  & AdamW & $1.0 \times 10^{-4}$ & $1.0 \times 10^{-3}$ \\
    4DOcc        & 50  & 2  & AdamW & $8.0 \times 10^{-5}$ & $8.0 \times 10^{-4}$ \\
    UnO          & 50  & 16 & AdamW & $8.0 \times 10^{-5}$ & $8.0 \times 10^{-4}$ \\
    \bottomrule
    \end{tabular}
    }
    \label{tab:pretraining}
\end{table}
\noindent \textbf{Pre-training Settings.} The detailed pre-training settings are shown in the table above. The learning rate (lr) follows a warmup-cosine schedule, increasing from ``lr start" to ``lr max" during the first 2\% of iterations (the warmup stage), and then follows a cosine schedule for the remaining iterations. All pre-training baseline methods maintain their original settings, while 4DOcc uses a very small batch size due to the high GPU memory consumption caused by the 0.1~m dense grid map.

\noindent \textbf{Fine-tuning Settings.} The Adam optimizer is used for all the fine-tuning; the learning rate follows a step-decay schedule, starting from $1.0 \times 10^{-4}$ and decreasing by a factor of $0.99$ after each iteration. For the nuScenes MOS experiments, the batch size is 4 for 5\%, 10\%, and 20\% data subsets, while the batch size for the 50\% data subset is 8. All methods are trained for over 400 epochs to ensure convergence. For SemanticKITTI cross-dataset transfer experiments, the batch size is 2 for all data subsets, and all methods are trained for over 300 epochs. For the nuScenes semantic segmentation experiment, all methods are trained for 300 epochs with a batch size of 4.

\section{Impact of Ego-Vehicle Points on IoU Metric}
\label{sec:iou}
As discussed in Sec.~4 of the main paper, the conventional IoU metric (denoted as IoU hereafter) includes ego-vehicle points, which inflates scores and masks actual perception performance on external moving objects. This issue is especially pronounced in the nuScenes dataset due to its high proportion of ego-vehicle points. To demonstrate this effect, we present nuScenes MOS results using IoU in Tab.~\ref{tab:nusc-ego}. The results show that IoU scores are inflated by 1.5-2x compared to the ego-exclusive metric $\text{IoU}_{\text{w/o}}$. Furthermore, the apparent IoU performance gains are deceptive. These improvements are not from enhanced perception of the external environment, but from the far simpler task of ego-vehicle segmentation. Therefore, relying on this metric risks severely misrepresenting a model's actual capability.

\begin{table}[ht]
\centering
\setlength\tabcolsep{4pt}
\scalebox{0.9}{
\begin{tabular}{c|c|ccc}
\toprule
\multirow{2}*{Data} & \multirow{2}*{Pretrain} & \multicolumn{3}{c}{Best $\textbf{Recall}_{\textbf{obj}}$} \\
\cmidrule(lr){3-5}
& & $\text{Recall}_{\text{obj}}$ & $\text{IoU}_{\text{w/o}}$ & $\text{IoU}$ \\
\midrule
\multirow{2}{*}{10\%}
                            & No & 24.98       &  36.44  &  68.89 \\
                            & \textbf{TOP}     & $\text{28.03}^{\text{\green{+3.04}}}$  
                                               &  $\text{36.95}^{\text{\green{+0.51}}}$  
                                               &  $\text{70.28}^{\text{\green{+1.39}}}$  \\
\midrule
\multirow{2}{*}{20\%}       & No & 25.59       &  44.25  &  63.68 \\
                            & \textbf{TOP}     & $\text{28.45}^{\text{\green{+2.86}}}$  
                                               & $\text{44.30}^{\text{\green{+0.05}}}$  
                                               & $\text{68.13}^{\text{\green{+4.45}}}$  \\
\bottomrule
\end{tabular}
}
\caption{nuScenes MOS results using the conventional IoU metric.}
\label{tab:nusc-ego}
\end{table}

\section{Potential Negative Social Impacts}
\label{sec:social}
While the proposed self-supervised method for MOS reduces annotation costs for autonomous systems, potential societal impacts should be considered. The technology could unintentionally prioritize detection accuracy for dominant object classes (e.g., vehicles) over vulnerable road users like cyclists/pedestrians if trained on imbalanced datasets, potentially compromising safety in edge cases. The temporal nature of our approach might propagate motion prediction errors in complex urban scenarios, leading to hazardous decisions by autonomous vehicles. Furthermore, while addressing sensor bias through our new metric helps, residual geographic/cultural biases in training data (e.g., urban vs rural environments) could limit global applicability. The method could also be repurposed for surveillance systems that infringe on privacy. To mitigate these risks, we recommend: (1) Rigorous testing across diverse operational domains. (2) Implementing fairness-aware data sampling strategies. (3) Establishing ethical guidelines for secondary applications.

\begin{figure*}[ht]
    \centering
    \includegraphics[width = 1\linewidth]{fig/nusc_vis_supp.png}
    \caption{Qualitative results of the nuScenes MOS.}
    \label{fig:nuscenes}
\end{figure*}

\begin{figure*}[ht]
    \centering
    \includegraphics[width = 1\linewidth]{fig/kitti_vis_supp.png}
    \caption{Qualitative results of the SemanticKITTI MOS.}
    \label{fig:kitti}
\end{figure*}

\end{document}